\documentclass[runningheads]{llncs}
\usepackage{graphicx}
\usepackage{tikz}
\usepackage{comment,verbatim}
\usepackage{amsmath,amssymb} % define this before the line numbering.
\usepackage{color,soul}
\usepackage[pagebackref,breaklinks,colorlinks]{hyperref}
\usepackage{multicol}
\usepackage{booktabs}
\usepackage{subcaption}

\usepackage[accsupp]{axessibility}  % Improves PDF readability for those with disabilities.

\begin{document}
% \linenumbers
\pagestyle{headings}
\mainmatter
%\def\ECCVSubNumber{100}  % Insert your submission number here

%%%%%%%%%%%%%%%%%%%%%%%%%%%%%%%%%%%%%%%%%%%%%%%

\title{ECCV 2024 W-CODA: 1st Workshop on Multimodal Perception and Comprehension of Corner Cases in Autonomous Driving} % Replace with your title

\titlerunning{ECCV 2024 W-CODA Workshop Report} 

\author{Kai Chen\inst{1}\thanks{Equal contribution. Contact: \email{kai.chen@connect.ust.hk}} \and
Ruiyuan Gao\inst{2}$^*$ \and
Lanqing Hong\inst{5}$^*$ \and
Hang Xu\inst{5} \and
Xu Jia\inst{3} \\
Holger Caesar\inst{4} \and
Dengxin Dai\inst{6} \and
Bingbing Liu\inst{5} \and
Dzmitry Tsishkou\inst{5} \\
Songcen Xu\inst{5} \and
Chunjing Xu\inst{7} \and
Qiang Xu\inst{2} \and
Huchuan Lu\inst{3} \and
Dit-Yan Yeung\inst{1}
}

\authorrunning{Chen et al.} 
\institute{
$^1$ Hong Kong University of Science and Technology \\
$^2$ The Chinese University of Hong Kong \\
$^3$ Dalian University of Technology \quad
$^4$ TU Delft \\
$^5$ Huawei Noah’s Ark Lab \quad
$^6$ Huawei Zurich Research Center \quad
$^7$ Huawei IAS BU \\
\url{https://coda-dataset.github.io/w-coda2024/}\\
\email{w-coda2024@googlegroups.com}
}

\maketitle

%%%%%%%%%%%%%%%%%%%%%%%%%%%%%%%%%%%%%%%%%%%%%%%

\vspace{-4mm}
\begin{abstract}

In this paper, we present details of the 1st W-CODA workshop, held in conjunction with the ECCV 2024. W-CODA aims to explore next-generation solutions for autonomous driving corner cases, empowered by state-of-the-art multimodal perception and comprehension techniques.
5 Speakers from both academia and industry are invited to share their latest progress and opinions.
We collect research papers and hold a dual-track challenge, including both corner case scene understanding and generation.
As the pioneering effort, we will continuously bridge the gap between frontier autonomous driving techniques and fully intelligent, reliable self-driving agents robust towards corner cases.

\vspace{-1.5mm}
\end{abstract}

%%%%%%%%%%%%%%%%%%%%%%%%%%%%%%%%%%%%%%%%%%%%%%%

\vspace{-5.5mm}
\section{Introduction}

This workshop aims to bridge the gap between state-of-the-art autonomous driving techniques and fully intelligent, reliable self-driving agents, particularly when confronted with corner cases~\cite{han2021soda10m,li2022coda,li2024automated}, rare but critical situations that challenge limits of reliable autonomous driving. 
The advent of Multimodal Large Language Models (MLLMs)~\cite{bai2025qwen2,chen2024emova,gou2023mixture,gpt4v_2} and AIGC~\cite{liu2023geomerasing,rombach2022high,wang2024detdiffusion} reveals remarkable abilities in multimodal perception and comprehension~\cite{gou2025perceptual,gou2024eyes,gou2025corrupted,wu2024unified} even under street scenes \cite{wen2023road}. 
However, leveraging MLLMs to tackle the nuanced challenges of self-driving remains an open field. 
This workshop seeks to foster innovative research work in \textit{multimodal perception and comprehension}, \textit{end-to-end driving systems}, and \textit{the application of advanced AIGC techniques to autonomous driving systems} (check Sec.~\ref{sec:papers}).
We conduct an international challenge consisting of two tracks, including \textit{corner case scene understanding} and \textit{corner case scene generation} (check Sec.~\ref{sec:challenge}).
This dual-track challenge is dedicated to advance reliability and interpretability of autonomous systems in both typical and extreme road corner cases.

%%%%%%%%%%%%%%%%%%%%%%%%%%%%%%%%%%%%%%%%%%%%%%%

\section{Call for Papers}\label{sec:papers}

\vspace{-2mm}
\subsubsection{Topics of Interest.} 
This workshop aims to foster innovative research in multimodal perception and comprehension of road corner cases, critical for advancing the next-generation industry-level self-driving solutions. 
Our focus encompasses a broad range of cutting-edge topics, including but not limited to:
\begin{itemize}
    \item Corner case mining and generation for autonomous driving.
    \item 3D object detection and scene understanding.
    \item Semantic occupancy prediction.
    \item Weakly supervised learning for 3D Lidar and 2D images.
    \item One/Few/Zero-shot learning for autonomous perception.
    \item End-to-end autonomous driving systems with Large Multimodal Models.
    \item Large Language Models techniques adoptable for self-driving systems.
    \item Safety/explainability/robustness for end-to-end autonomous driving.
    \item Domain adaptation and generalization for end-to-end autonomous driving.
\end{itemize}

\vspace{-3mm}
\subsubsection{Submission Tracks.} 
All submissions must be anonymous, and the reviewing procedure is double-blind. We encourage two types of submissions:
\begin{itemize}
    \item \textbf{Full workshop papers} not previously published or accepted for any official publications in a substantially similar form in any peer-reviewed venues, including journals, conferences, or workshops. Papers are limited to 14 pages, including both figures and tables, in the ECCV format, with additional pages containing only cited references allowed. Accepted papers will be part of the official ECCV proceedings.
    
    \item \textbf{Extended abstract papers} not previously published or accepted for any publications in substantially similar form in any other peer-reviewed venues, including journals, conferences, or workshops. Papers are limited to 4 pages and will NOT be included in official ECCV proceedings (non-archival), which are permitted for resubmission to later conferences.
\end{itemize}

\vspace{-3mm}
\subsubsection{Accepted papers (full track)} are collected as part of the official Computer Vision – ECCV 2024 Workshops proceedings (part \uppercase\expandafter{\romannumeral7})\footnote{\url{https://link.springer.com/book/10.1007/978-3-031-91767-7}}, including:
\begin{itemize}
    \item \textbf{Paper ID 2:} \href{https://openreview.net/forum?id=Xzb1QlJ98b}{AnoVox: A Benchmark for Multimodal Anomaly Detection in Autonomous Driving} - \textit{Daniel Bogdoll, Iramm Hamdard, Lukas Roessler, Felix Geisler, Muhammed Bayram, Felix Wang, Jan Imhof, Miguel de Campos, Anushervon Tabarov, Yitian Yang, Martin Gontscharow, Hanno Gottschalk, J. Marius Zoellner}
    \item \textbf{Paper ID 3:} \href{https://openreview.net/forum?id=nFA8MWo9Ks}{On Camera and LiDAR Positions in End-to-End Autonomous Driving} - \textit{Malte Stelzer, Jan Pirklbauer, Jan Bickerdt, Volker Patricio Schomerus, Jan Piewek, Thorsten Bagdonat, Tim Fingscheidt}
    \item \textbf{Paper ID 4:} \href{https://openreview.net/forum?id=5nglB4rkux}{ProGBA: Prompt Guided Bayesian Augmentation for Zero-shot Domain Adaptation} - \textit{Jian Zou, Guanglei Yang, Tao Luo, Chun-Mei Feng, Wangmeng Zuo}
    \item \textbf{Paper ID 7:} \href{https://openreview.net/forum?id=dJqcdUgEdw}{ReGentS: Real-World Safety-Critical Driving Scenario Generation Made Stable} - \textit{Yuan Yin, Pegah KHAYATAN, Eloi Zablocki, Alexandre Boulch, Matthieu Cord}
    \item \textbf{Paper ID 8:} \href{https://openreview.net/forum?id=SyqqPooxxm}{Loop Mining Large-Scale Unlabeled Data for Corner Case Detection in Autonomous Driving} - \textit{Jiawei Zhao, Yiting Duan, Jinming Su, yangwangwang, Tingyi Guo, Xingyue Chen, Junfeng Luo}
    \item \textbf{Paper ID 9:} \href{https://openreview.net/forum?id=lVMKJxsIdC}{HumanSim: Human-Like Multi-Agent Novel Driving Simulation for Corner Case Generation} - \textit{Lingfeng Zhou, Mohan Jiang, Dequan Wang}
    \item \textbf{Paper ID 11:} \href{https://openreview.net/forum?id=zLWJR53KxC}{Talk to Parallel LiDARs: A Human-LiDAR Interaction Method Based on 3D Visual Grounding} - \textit{Yuhang Liu, Boyi Sun, Yishuo Wang, Jing Yang, Xingxia Wang, Fei-Yue Wang}
    \item \textbf{Paper ID 16:} \href{https://openreview.net/forum?id=ygF6aFhdxC}{RoSA Dataset: Road construct zone Segmentation for Autonomous Driving} - \textit{JINWOO KIM, Kyounghwan An, Donghwan Lee}
    \item \textbf{Paper ID 17:} \href{https://openreview.net/forum?id=ad6Q54fHMr}{A Multimodal Hybrid Late-Cascade Fusion Network for Enhanced 3D Object Detection} - \textit{Carlo Sgaravatti, Roberto Basla, Riccardo Pieroni, Matteo Corno, Sergio M. Savaresi, Luca Magri, Giacomo Boracchi}
\end{itemize}

\vspace{-6mm}
\subsubsection{Accepted papers (abstract track)} include:
\begin{itemize}
    \item \textbf{Paper ID 2:} \href{https://openreview.net/forum?id=dvS8rYvXDh}{Challenge report: Track 2 of Multimodal Perception and Comprehension of Corner Cases in Autonomous Driving} - \textit{Zhiying Du, Zhen Xing}
    \item \textbf{Paper ID 3:} \href{https://openreview.net/forum?id=wP3GZbi1mN}{DreamForge: Motion-Aware Autoregressive Video Generation for Multi-View Driving Scenes} - \textit{Jianbiao Mei, Yukai Ma, Xuemeng Yang, Licheng Wen, Tiantian Wei, Min Dou, Botian Shi, Yong Liu}
    \item \textbf{Paper ID 4:} \href{https://openreview.net/forum?id=0r6HZXwvra}{Two-Stage LVLM system: 1st Place Solution for ECCV 2024 Corner Case Scene Understanding Challenge} - \textit{Ying Xue, Haiming Zhang, Yiyao Zhu, Wending Zhou, Shuguang Cui, Zhen Li}
    \item \textbf{Paper ID 5:} \href{https://openreview.net/forum?id=LXZO1nGI0d}{NexusAD: Exploring the Nexus for Multimodal Perception and Comprehension of Corner Cases in Autonomous Driving} - \textit{Mengjingcheng Mo, Jingxin Wang, Like Wang, Haosheng Chen, Changjun Gu, Jiaxu Leng, Xinbo Gao}
    \item \textbf{Paper ID 6:} \href{https://openreview.net/forum?id=QE5IUwmBuL}{DiVE: DiT-based Video Generation with Enhanced Control} - \textit{Junpeng Jiang, Gangyi Hong, Lijun Zhou, Enhui Ma, Hengtong Hu, Xia Zhou, Jie Xiang, Fan Liu, Kaicheng Yu, Haiyang Sun, Kun Zhan, Peng Jia, Miao Zhang}
    \item \textbf{Paper ID 7:} \href{https://openreview.net/forum?id=BGwuAnKSMu}{From Regional to General: A Vision-Language Model-Based Framework for Corner Cases Comprehension in Autonomous Driving} - \textit{XU HAN, Yehua Huang, SongTang, Xiaowen Chu}
    \item \textbf{Paper ID 8:} \href{https://openreview.net/forum?id=2INNKtsrYT}{A Lightweight Vision-Language Model Pipeline for Corner-Case Scene Understanding in Autonomous Driving} - \textit{Ying Cheng, Min-Hung Chen, Shang-Hong Lai}
    \item \textbf{Paper ID 9:} \href{https://openreview.net/forum?id=sZkaCgx0CI}{Iterative Finetuning VLM with Retrieval-augmented Synthetic Datasets Technical Reports for W-CODA Challenge Track-1 from Team OpenDriver} - \textit{Zihao Wang, Xueyi Li}
    \item \textbf{Paper ID 11:} \href{https://openreview.net/forum?id=Xx00r8A67P}{FORTRESS: Feature Optimization and Robustness Techniques for 3D Object Detection Systems} - \textit{Caixin Kang, Xinning Zhou, Chengyang Ying, Wentao Shang, Xingxing Wei, Yinpeng Dong, Hang Su}
    \item \textbf{Paper ID 12:} \href{https://openreview.net/forum?id=aU0ocxMy0u}{Adversarial Policy Generation in Automated Parking} - \textit{Alessandro Pighetti, Francesco Bellotti, Riccardo Berta, Andrea Cavallaro, Luca Lazzaroni, Changjae Oh}
\end{itemize}

%%%%%%%%%%%%%%%%%%%%%%%%%%%%%%%%%%%%%%%%%%%%%%%

\section{The W-CODA Challenge}\label{sec:challenge}

This challenge aims to establish a comprehensive, industry-level benchmark for evaluating the generalization and robustness of the multimodal perception and understanding models within real-world autonomous driving applications.

Scheduled for release before June 15th, 2024, we provide participants opportunities to showcase their techniques. The most successful and innovative entries will be highlighted at our workshop, with their creators receiving recognition and awards. Prizes include a 1,000 USD cash reward for the top performers in each track, with the second and third receiving 800 USD and 600 USD, respectively.

%%%%%%%%%%%%%%%%%%%%%%%%%%%%%%%%%%%%%%%%%%%%%%%

\vspace{-2mm}
\subsection{Track 1: Corner Case Scene Understanding}

\subsubsection{Overview.}
This track\footnote{\url{https://coda-dataset.github.io/w-coda2024/track1/}} is dedicated to enhancing the multimodal perception and comprehension abilities of MLLMs for autonomous driving, focusing on the global scene understanding, local regional reasoning, and actionable navigation. 
With our CODA-LM~\cite{li2024automated} dataset constructed based on CODA~\cite{li2022coda} via collecting LLM-synthetic data~\cite{chen2023gaining,liu2024mixture} followed by manual inspection, which comprises about 10K images with corresponding textual annotations covering global driving scenarios, detailed analyses of corner cases, and driving suggestions, this competition seeks to promote the development of more reliable and interpretable autonomous driving agents.

\vspace{-2mm}
\subsubsection{Task Description.}
CODA-LM contains three tasks to promote the development of more reliable and interpretable autonomous driving agents.

\begin{itemize}
    \item \textbf{General Perception.} 
    The foundational aspect of this task lies in the comprehensive understanding of key entities within driving scenes, including the appearance, location, and the reasons they influence driving. Given a first-view road driving scene, MLLMs are required to describe all potential road obstacles in the traffic scenes and explain why they would affect the driving decisions. We primarily focus on seven categories of obstacles, including vehicles, vulnerable road users (VRUs), traffic cones, traffic lights, traffic signs, barriers, and miscellaneous.
    \item \textbf{Region Perception.} 
    This task measures the MLLMs' capability to understand corner objects when provided with specific bounding boxes as visual prompts, followed by describing these objects and explaining why they might affect self-driving behavior.    
    Note that for region perception, there are no constraints on how to utilize bounding boxes in MLLMs. Any possible encoding leading to better performance is feasible. 
    We provide an example by visualizing the bounding boxes with red rectangles, as in the original CODA-LM paper (\textit{c.f.}, Fig. 2 in \cite{li2024automated}) for reference.
    \item \textbf{Driving Suggestions.} 
    This task aims to evaluate the MLLMs' capability in formulating actionable driving suggestions, closely related to the planning process of autonomous driving, requiring MLLMs to provide optimal driving suggestions for the ego car after correctly perceiving the general and regional aspects of the current driving environment.
\end{itemize}

\vspace{-3mm}
\subsubsection{Results.}
\begin{figure}[t]
  \centering
  \begin{subfigure}{0.49\linewidth}
    \includegraphics[width=1.0\linewidth]{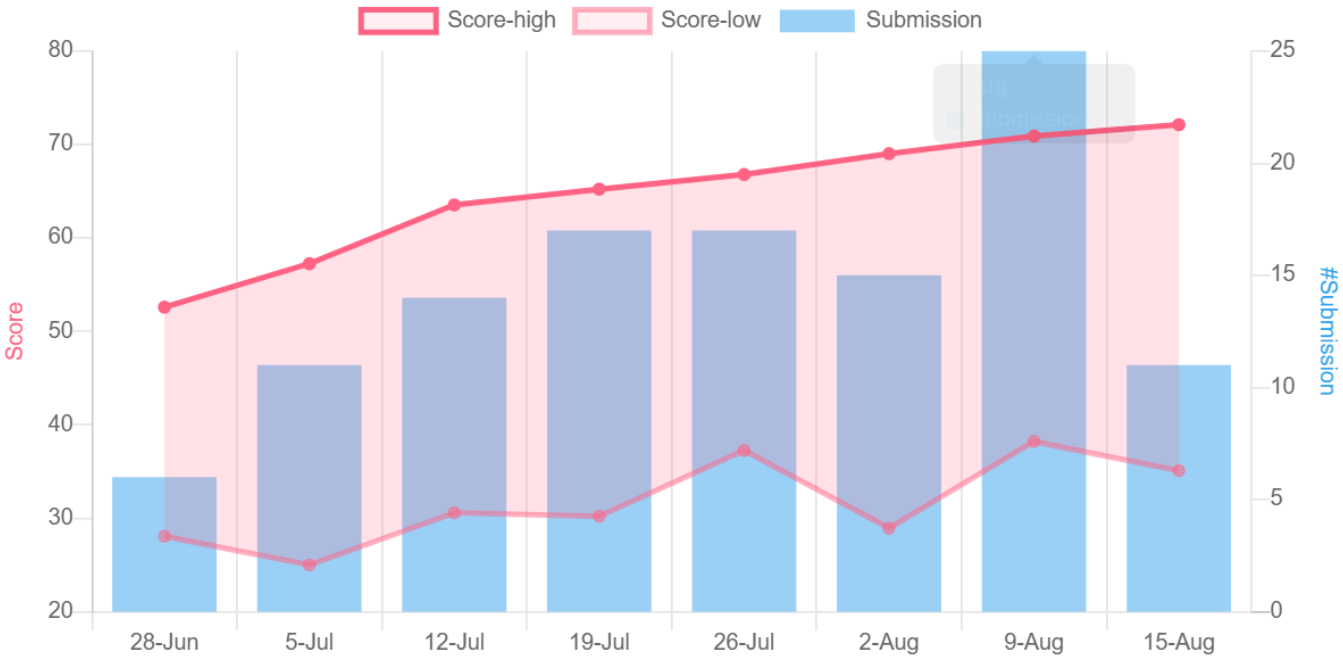}
    \caption{\textbf{Progress curve of Track 1.}}
    \label{fig:track1}
  \end{subfigure}
  \hfill
  \begin{subfigure}{0.49\linewidth}
    \includegraphics[width=1.0\linewidth]{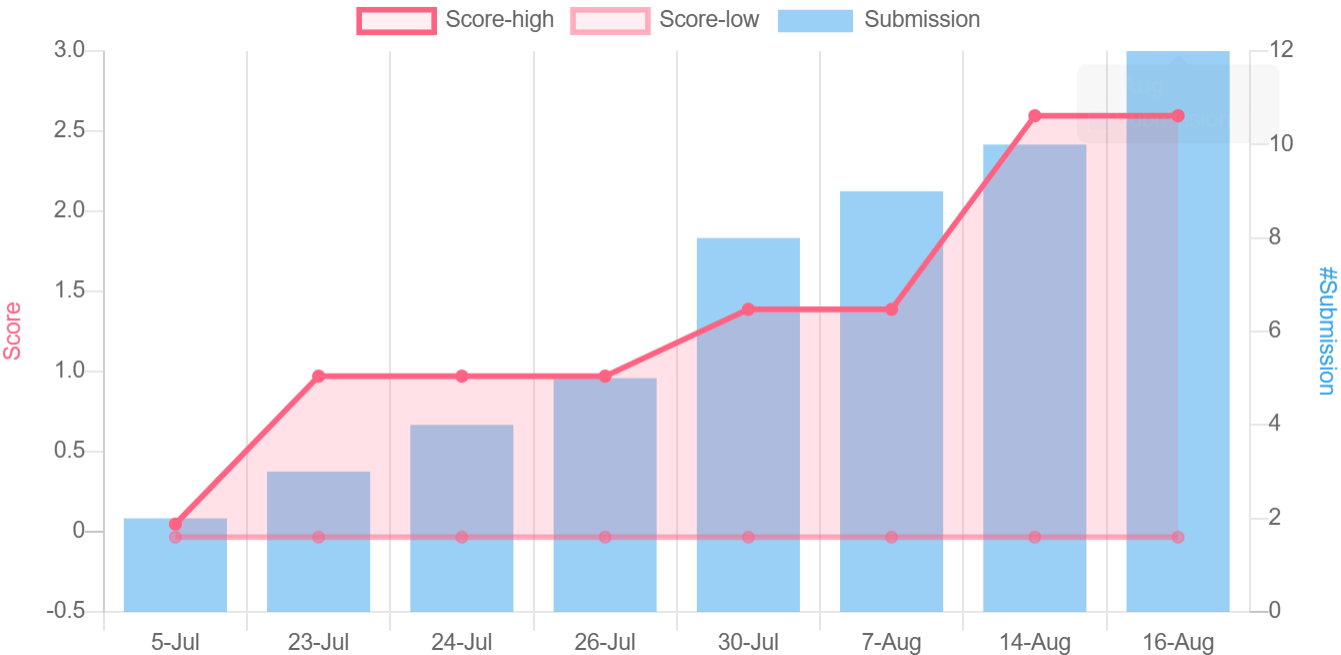}
     \caption{\textbf{Progress curve of Track 2.}}
    \label{fig:track2}
  \end{subfigure}
  \vspace{-2mm}
  \caption{\textbf{Progress curves of the dual-track W-CODA challenge.}
  }
  \vspace{-3mm}
  \label{fig:challenge}
\end{figure}

As shown in the progress curve in Fig.~\ref{fig:track1}, we receive 116 submissions from 56 registered teams spanning 100+ individuals from 30+ institutions.
The best-performing winner obtains remarkable improvements over organizer baselines (\textit{i.e.}, \textbf{+156.02\%} over LLaVA-1.5~\cite{liu2024improved} and \textbf{+13.36\%} over CODA-VLM \cite{li2024automated} for the Final Score).
We collect winners' solutions and reports in the following:
\begin{itemize}
    \item \textbf{Rank \#1:} \href{https://openreview.net/forum?id=0r6HZXwvra}{Team llmforad} (Ying Xue, Haiming Zhang, Yiyao Zhu, Wending Zhou, Shuguang Cui, Zhen Li) - \textit{The Chinese University of Hong Kong (Shenzhen), Hong Kong University of Science and Technology}
    \item \textbf{Rank \#2:} \href{https://openreview.net/forum?id=sZkaCgx0CI}{Team OpenDriver} (Zihao Wang, Xueyi Li) - \textit{Peking University}
    \item \textbf{Rank \#3:} \href{https://openreview.net/forum?id=LXZO1nGI0d}{Team NexusAD} (Mengjingcheng Mo, Jingxin Wang, Like Wang, Haosheng Chen, Changjun Gu, Jiaxu Leng, Xinbo Gao) - \textit{Chongqing University of Posts and Telecommunications}
\end{itemize}

%%%%%%%%%%%%%%%%%%%%%%%%%%%%%%%%%%%%%%%%%%%%%%%

%%%%%%%%%%%%%%%%%%%%%%%%%%%%%%%%%%%%%%%%%%%%%%%

\vspace{-3mm}
\subsection{Track 2: Corner Case Scene Generation}

\subsubsection{Overview.}
This track\footnote{\url{https://coda-dataset.github.io/w-coda2024/track2/}} focuses on improving geometric-controllability~\cite{chen2023integrating,gao2023magicdrive,li2023trackdiffusion} of diffusion models to generate high-quality multi-view street scene videos which are consistent with 3D geometric scene descriptors (\textit{e.g.}, Bird's Eye View (BEV) maps and 3D LiDAR bounding boxes).
Building on MagicDrive~\cite{gao2024magicdrive3d,gao2024magicdrivedit} for controllable multi-view video generation, this track aims to advance the scene generation and world modeling for autonomous driving, ensuring better consistency, higher resolution, and longer duration.

\vspace{-3mm}
\subsubsection{Task Description.}
In this track, participants should train a controllable multi-view video generation model for the street scenes. The generated videos should accurately reflect control signals from the BEV road maps, 3D bounding boxes, and textual descriptions of weather and time-of-day. 
We will evaluate the generated videos from generation quality and controllability.
Furthermore, we have provided a 16-frame organizer baseline model with configs and pre-trained weights based on MagicDrive~\cite{gao2023magicdrive} to help the participants better started.

\vspace{-3mm}
\subsubsection{Results.}
As shown in the progress curve in Fig.~\ref{fig:track2}, we receive 12 submissions from 12 registered teams spanning 40+ individuals from 20+ institutions.
The best-performing winner obtains remarkable improvements over organizer baselines (\textit{i.e.}, \textbf{+56.63\%} for FVD, \textbf{+106.97\%} for mAP, and \textbf{+96.02\%} for mIoU over MagicDrive~\cite{gao2023magicdrive}).
We collect winners' solutions and reports in the following:
\begin{itemize}
    \item \textbf{Rank \#1:} \href{https://openreview.net/forum?id=QE5IUwmBuL}{Team LiAuto-AD} (Junpeng Jiang, Gangyi Hong, Lijun Zhou, Enhui Ma, Hengtong Hu, xia zhou, Jie Xiang, Fan Liu, Kaicheng Yu, Haiyang Sun, Kun Zhan, Peng Jia, Miao Zhang) - \textit{Harbin Institute of Technology (Shenzhen), Li Auto Inc., Tsinghua University, Westlake University, National University of Singapore}
    \item \textbf{Rank \#2:} \href{https://openreview.net/forum?id=wP3GZbi1mN}{Team DreamForge} (Jianbiao Mei, Yukai Ma, Xuemeng Yang, Licheng Wen, Tiantian Wei, Min Dou, Botian Shi, Yong Liu) - \textit{Zhejiang University, Shanghai Artificial Intelligence Laboratory, Technical University of Munich}
    \item \textbf{Rank \#3:} \href{https://openreview.net/forum?id=dvS8rYvXDh}{Team Seven} (Zhiying Du, Zhen Xing) - \textit{Shanghai Key Lab of Intell. Info. Processing, School of CS, Fudan University}
\end{itemize}

%%%%%%%%%%%%%%%%%%%%%%%%%%%%%%%%%%%%%%%%%%%%%%%

\vspace{-3mm}
\section{Program Outline}

\subsubsection{Overview.} 
We plan a half-day event for this workshop, which will be roughly split among invited speakers and challenge winners.
This workshop includes peer-reviewed papers, which will be published in the proceedings (CVF only). 
We plan to hold 5 invited talks, and 2 sessions of oral presentations given by challenge winners.
We will need a digital projector (for presenters) and a flipchart with pens (for publicly recording small-group critiques).

\vspace{-3mm}
\subsubsection{Program Schedule.}
\begin{itemize}
    \item 09:00-09:05 Opening remarks, welcome, and challenge summary.
    \item 09:05-09:35 Invited talk 1: Prof. Antonio M. López (UAB). \\
    \textbf{Vision-based End-to-end Driving by Imitation Learning}
    \item 09:35-10:05 Invited talk 2: Prof. Hongyang Li (HKU). \\
    \textbf{Reasoning Multi-Agent Behavioral Topology for Interactive Autonomous Driving}
    \item 10:05-10:35 Invited talk 3: Prof. Andreas Geiger (University of Tübingen). \\
    \textbf{Simulating and Benchmarking Self-Driving Cars}
    \item 10:35-11:05 Poster Session and Coffee Break.
    \item 11:05-11:35 Invited talk 4: Dr. Lorenzo Bertoni (Wayve). \\
    \textbf{Long-tail Scenario Generation for Autonomous Driving with World Models}
    \item 11:35-12:05 Invited talk 5: Dr. Chufeng Tang (Huawei). \\
    \textbf{Industrial Talk on Autonomous Driving}
    \item 12:05-12:15 Challenge Summary \& Awards.
    \item 12:15-12:30 Oral talk 1: Winners of Track 1.
    \item 12:30-12:45 Oral talk 2: Winners of Track 2.
    \item 12:45-13:00 Summary \& Future Plans.
\end{itemize}

\vspace{-6mm}
\subsubsection{Important Dates (AoE Time, UTC-12).}
\begin{itemize}
    \item Challenge Open to Public: June 15, 2024.
    \item Challenge Submission Deadline: Aug 15, 2024 11:59 PM.
    \item Challenge Notification to Winner: Sep 1, 2024.
    \item Full Paper Submission Deadline: Aug 1, 2024 11:59 PM.
    \item Full Paper Notification to Authors: Aug 10, 2024.
    \item Full Paper Camera Ready Deadline: Sep 18, 2024 11:59 PM.
    \item Abstract Paper Submission Deadline: Sep 1, 2024 11:59 PM.
    \item Abstract Paper Notification to Authors: Sep 7, 2024.
    \item Abstract Paper Camera Ready Deadline: Sep 10, 2024 11:59 PM.
\end{itemize}

\vspace{-6mm}
\subsubsection{Invited Speakers.}
\begin{itemize}
    \item \textbf{Prof. Antonio M. López (UAB)} \\
    Antonio M. López is a Professor at the Universitat Autònoma de Barcelona. He has a long trajectory carrying research at the intersection of computer vision, simulation, machine learning, driver assistance, and autonomous driving. Antonio has been deeply involved in the creation of the SYNTHIA and UrbanSyn datasets and the CARLA open-source simulator, all created for democratizing autonomous driving research. Antonio’s team was a pioneer in synth-to-real domain adaptation in the late 2010s. Antonio’s team and colleagues also put the focus on vision-based end-to-end autonomous driving powered by deep imitation learning. Antonio is actively working hand-in-hand with industry partners to bring state-of-the-art techniques to the field of autonomous driving.
    
    \item \textbf{Prof. Hongyang Li (HKU)} \\
    Hongyang Li is an Assistant Professor at the University of Hong Kong and a Research Scientist at OpenDriveLab, Shanghai AI Lab. His research focuses on self-driving and embodied AI. He led the end-to-end autonomous driving project, UniAD, and won the CVPR 2023 Best Paper Award. UniAD has a large impact both in academia and industry, including the recent rollout to customers by Tesla in FSD V12. He served as Area Chair for CVPR 2023, 2024, NeurIPS 2023 (Notable AC), 2024, ACM MM 2024, and referee for Nature Communications. He will serve as Workshop Chair for CVPR 2026. He is the Working Group Chair for IEEE Standards under the Vehicular Technology Society and a Senior Member of IEEE.
    
    \item \textbf{Prof. Andreas Geiger (University of Tübingen)} \\
    Andreas Geiger is a Professor at the University of Tübingen and the Tübingen AI Center. Previously, he was a visiting professor at ETH Zürich and a group leader at the Max Planck Institute for Intelligent Systems. He studied at KIT, EPFL, and MIT, and received his PhD degree in 2013 from KIT. His research focuses on the intersection of computer vision, machine learning, and robotics. His work has received the Longuet-Higgins Prize, the Mark Everingham Prize, the IEEE PAMI Young Investigator Award, the Heinz Maier-Leibnitz Prize, and the German Pattern Recognition Award. In 2013 and 2021, he received the CVPR best paper and best paper runner-up awards. He also received the best paper award at GCPR 2015 and 3DV 2015, as well as the best student paper award at 3DV 2017. In 2019, he was awarded an ERC starting grant. He is an ELLIS fellow and coordinates the ELLIS PhD and PostDoc program. He maintains the KITTI and KITTI-360 benchmarks.
    
    \item \textbf{Dr. Lorenzo Bertoni (Wayve)} \\
    Lorenzo Bertoni is a Senior Applied Scientist at Wayve, where he works at the intersection of generative AI and autonomous driving. He earned his Ph.D. with distinction from EPFL in 2022, focusing on 3D perception for autonomous driving. His research emphasized improving safety for vulnerable road users and addressing corner cases. Lorenzo has published at leading computer vision and robotics conferences such as CVPR, ICCV, and ICRA. His work has been featured in EPFL news, various Swiss newspapers, and on the BBC 4Tech Show.
    
    \item \textbf{Dr. Chufeng Tang (Huawei)} \\
    Chufeng Tang is now a researcher in the IAS BU of Huawei. His research interests primarily lie in computer vision and autonomous driving. He received his Ph.D. degree from Tsinghua University in 2023 and his B.E. degree from Huazhong University of Science and Technology in 2018.
\end{itemize}

%%%%%%%%%%%%%%%%%%%%%%%%%%%%%%%%%%%%%%%%%%%%%%%

\vspace{-5mm}
\section{Relations to Previous Workshops}

Autonomous driving has been a long-standing workshop topic for the community, focusing on general perception, end-to-end driving, representation learning~\cite{chen2021multisiam,chen2023mixed,liu2022task,zhili2023task}, and data simulation for autonomous driving, and this workshop differs from previous events in the following two perspectives: 

1) We emphasize the importance of the \textbf{corner cases} for reliable autonomous driving, and challenge the recognition and adaptation for novel and unseen road scenes of autonomous systems. 
2) We feature the usage of \textbf{MLLMs and AIGC} empowered by visual generative models, pursuing interactable and interpretable driving agents for the next-generation industry-level self-driving solutions.

The most relevant to W-CODA is the 2nd SSLAD Workshop at ECCV 2022\footnote{\url{https://sslad2022.github.io/}}, also held by our core organization team, and the Workshop on Foundation Models for Autonomous Systems (FMAS) at CVPR 2024\footnote{\url{https://opendrivelab.com/cvpr2024/workshop/}}, held by OpenDriveLab.
The 2nd SSLAD Workshop, for the first time, directly evaluates the performance of autonomous systems on corner cases, which, however, is limited to 2D detection, while W-CODA further extends to multimodal perception and reasoning, as in Sec.~\ref{sec:challenge}.
Instead, the FMAS Workshop is the first to evaluate the capabilities of LLMs for autonomous driving, which, however, is still limited to normal street scenes in the nuScenes~\cite{caesar2020nuscenes} dataset, while ours concentrates on the critical road corner cases, and collects the challenging CODA-LM~\cite{li2024automated} dataset, which forms the peerless advantage of our workshop.

%%%%%%%%%%%%%%%%%%%%%%%%%%%%%%%%%%%%%%%%%%%%%%%

\section{Relevance to the Community}

This workshop aims at bridging the gap among the state-of-the-art autonomous driving technologies and the vision of fully intelligent, reliable self-driving agents, underscoring the critical importance of robust visual perception, adept decision-making, and precise planning controls.
Traditional rule-based control systems fall behind in addressing the unpredictability and complexity of real-life driving scenarios, particularly when confronted with the corner cases, rare but critical situations that challenge the limits of reliable automated systems~\cite{breitenstein2021corner,li2022coda}.

The advent of MLLMs has introduced a paradigm shift, revealing unprecedented capabilities in both visual perception and reasoning, even under dynamic street scenes~\cite{wen2023road}, hinting at the potential for enhancing autonomous driving systems' understanding and analytical capabilities, especially in navigating the intricacies of corner cases.
Nonetheless, fully leveraging LLMs to tackle the nuanced challenges of autonomous driving, especially for the 3D perception, a cornerstone for navigation, motion prediction, and collision avoidance, still remains an open field, ripe for throughout exploration and innovation. 

The relevance of this workshop to the computer vision community is profound and multifaceted. 
It seeks to foster innovative research and development in multimodal perception and comprehension, end-to-end driving systems, and the application of advanced LLM and AIGC techniques to autonomous driving.
The focus on corner case mining and generation, along with the efforts to improve robustness and generalization in autonomous systems, is particularly pertinent, which not only aligns with but also pushes boundaries of current research in computer vision, as evidenced by the rapid growth of related publications in top-tier conferences and journals such as CVPR, ICCV, ECCV, and T-PAMI.

The viability of this workshop is reinforced by its alignment with academic and industry efforts to tackle the intricate challenges of corner cases. 
Designed to foster interdisciplinary dialogue and collaboration, it is set to accelerate the progress towards next-generation industry-level autonomous systems, aiming to bridge the gap between self-driving research and large-scale real-world applications, a topic currently underrepresented in major conferences. 
With confirmed participation from the renowned professors and researchers ready to contribute through keynote speeches and challenges, this workshop promises to draw a varied audience of active researchers, practitioners, and industry professionals. 
The collective dedication for advanced autonomous driving and computer vision assures a significant impact, making it not only viable but also poised for success.

As detailed in Sec.~\ref{sec:challenge}, this workshop plans to conduct a dual-track challenge designed to advance the reliability, interpretability, and safety of self-driving by emphasizing robustness in multimodal perception and critical problem-solving in both typical and extreme driving conditions (\textit{i.e.}, corner cases).

In this half-day workshop, we will host the regular paper presentation, invited talks, and technical challenges, as well as limitations and future directions for computer vision and autonomous driving. 

%%%%%%%%%%%%%%%%%%%%%%%%%%%%%%%%%%%%%%%%%%%%%%%

\section{Organization Team Experience}

Our organization team boasts a profound background for both academic and industrial autonomous driving. 
Prof. Huchuan Lu, IEEE Fellow from DLUT, has published more than 100 research papers on object tracking for autonomous driving with more than 40,000 citations. 
Dr. Dengxin Dai, Director of Huawei Zurich Research Center, and Dr. Bingbing Liu, Technical Expert of Huawei Noah’s Ark Lab, have numerous patents on self-driving techniques and are deeply involved in the construction and promotion of Huawei Advanced Driving System (ADS)~\footnote{\url{https://www.huawei.com/en/
media-center/our-value/advanced-driving-system}}, possessing a significant lead over global competitors.

Moreover, our team is also equipped with extensive experience in organizing workshops on autonomous driving in top-tier conferences.  
Dr. Dengxin Dai organized the \textit{Autonomous Driving Workshop} at ICCV 2019, the \textit{Synthetic Data for Autonomous Systems Workshop} at CVPR 2023 and the \textit{BRAVO: Robustness and Reliability of Autonomous Vehicles in the Open-world Workshop} at ICCV 2023. 
Dr. Lanqing Hong and Dr. Kai Chen organized the \textit{Self-supervised Learning for Next-Generation Industry-level Autonomous Driving (SSLAD) Workshop} (1st at ICCV 2021 and 2nd at ECCV 2022). 
Prof. Jia Xu organized the \textit{Continual Learning in Computer Vision Workshop} at CVPR 2021.
Thus, we are confident in our expertise to well organize this workshop at ECCV 2024.

%%%%%%%%%%%%%%%%%%%%%%%%%%%%%%%%%%%%%%%%%%%%%%%

\section{Diversity}

The researchers in the list of organizers and speakers come from various countries, include both males and females, work in either academia or industry, and work on various AI topics (including CV, NLP, and Robotics), which demonstrates significant diversity and inclusion of the proposed workshop.

%%%%%%%%%%%%%%%%%%%%%%%%%%%%%%%%%%%%%%%%%%%%%%%

\section{Anticipated Target Audience}

Considering the heightened attention and autonomous driving technologies and the importance of corner cases, our anticipated target audiences embrace global talents working on autonomous driving, including scholars and students from universities and research institutions, and researchers and engineers from intelligent car companies.

\bibliographystyle{splncs04}
\bibliography{egbib}

\begin{thebibliography}{10}
\providecommand{\url}[1]{\texttt{#1}}
\providecommand{\urlprefix}{URL }
\providecommand{\doi}[1]{https://doi.org/#1}

\bibitem{bai2025qwen2}
Bai, S., Chen, K., Liu, X., Wang, J., Ge, W., Song, S., Dang, K., Wang, P., Wang, S., Tang, J., et~al.: Qwen2. 5-vl technical report. arXiv preprint arXiv:2502.13923  (2025)

\bibitem{breitenstein2021corner}
Breitenstein, J., Term{\"o}hlen, J.A., Lipinski, D., Fingscheidt, T.: Corner cases for visual perception in automated driving: Some guidance on detection approaches. arXiv preprint arXiv:2102.05897  (2021)

\bibitem{caesar2020nuscenes}
Caesar, H., Bankiti, V., Lang, A.H., Vora, S., Liong, V.E., Xu, Q., Krishnan, A., Pan, Y., Baldan, G., Beijbom, O.: nuscenes: A multimodal dataset for autonomous driving. In: CVPR (2020)

\bibitem{chen2024emova}
Chen, K., Gou, Y., Huang, R., Liu, Z., Tan, D., Xu, J., Wang, C., Zhu, Y., Zeng, Y., Yang, K., et~al.: Emova: Empowering language models to see, hear and speak with vivid emotions. arXiv preprint arXiv:2409.18042  (2024)

\bibitem{chen2021multisiam}
Chen, K., Hong, L., Xu, H., Li, Z., Yeung, D.Y.: Multisiam: Self-supervised multi-instance siamese representation learning for autonomous driving. In: ICCV (2021)

\bibitem{chen2023mixed}
Chen, K., Liu, Z., Hong, L., Xu, H., Li, Z., Yeung, D.Y.: Mixed autoencoder for self-supervised visual representation learning. In: CVPR (2023)

\bibitem{chen2023gaining}
Chen, K., Wang, C., Yang, K., Han, J., Hong, L., Mi, F., Xu, H., Liu, Z., Huang, W., Li, Z., Yeung, D.Y., Shang, L., Jiang, X., Liu, Q.: Gaining wisdom from setbacks: Aligning large language models via mistake analysis. arXiv preprint arXiv:2310.10477  (2023)

\bibitem{chen2023integrating}
Chen, K., Xie, E., Chen, Z., Hong, L., Li, Z., Yeung, D.Y.: Integrating geometric control into text-to-image diffusion models for high-quality detection data generation via text prompt. arXiv preprint arXiv:2306.04607  (2023)

\bibitem{gao2024magicdrive3d}
Gao, R., Chen, K., Li, Z., Hong, L., Li, Z., Xu, Q.: Magicdrive3d: Controllable 3d generation for any-view rendering in street scenes. arXiv preprint arXiv:2405.14475  (2024)

\bibitem{gao2024magicdrivedit}
Gao, R., Chen, K., Xiao, B., Hong, L., Li, Z., Xu, Q.: Magicdrivedit: High-resolution long video generation for autonomous driving with adaptive control. arXiv preprint arXiv:2411.13807  (2024)

\bibitem{gao2023magicdrive}
Gao, R., Chen, K., Xie, E., Hong, L., Li, Z., Yeung, D.Y., Xu, Q.: Magicdrive: Street view generation with diverse 3d geometry control. arXiv preprint arXiv:2310.02601  (2023)

\bibitem{gou2025perceptual}
Gou, Y., Chen, K., Liu, Z., Hong, L., Jin, X., Li, Z., Kwok, J.T., Zhang, Y.: Perceptual decoupling for scalable multi-modal reasoning via reward-optimized captioning. arXiv preprint arXiv:2506.04559  (2025)

\bibitem{gou2024eyes}
Gou, Y., Chen, K., Liu, Z., Hong, L., Xu, H., Li, Z., Yeung, D.Y., Kwok, J.T., Zhang, Y.: Eyes closed, safety on: Protecting multimodal llms via image-to-text transformation. arXiv preprint arXiv:2403.09572  (2024)

\bibitem{gou2023mixture}
Gou, Y., Liu, Z., Chen, K., Hong, L., Xu, H., Li, A., Yeung, D.Y., Kwok, J.T., Zhang, Y.: Mixture of cluster-conditional lora experts for vision-language instruction tuning. arXiv preprint arXiv:2312.12379  (2023)

\bibitem{gou2025corrupted}
Gou, Y., Yang, H., Liu, Z., Chen, K., Zeng, Y., Hong, L., Li, Z., Liu, Q., Kwok, J.T., Zhang, Y.: Corrupted but not broken: Rethinking the impact of corrupted data in visual instruction tuning. arXiv preprint arXiv:2502.12635  (2025)

\bibitem{han2021soda10m}
Han, J., Liang, X., Xu, H., Chen, K., Hong, L., Ye, C., Zhang, W., Li, Z., Liang, X., Xu, C.: Soda10m: Towards large-scale object detection benchmark for autonomous driving. arXiv preprint arXiv:2106.11118  (2021)

\bibitem{li2022coda}
Li, K., Chen, K., Wang, H., Hong, L., Ye, C., Han, J., Chen, Y., Zhang, W., Xu, C., Yeung, D.Y., et~al.: Coda: A real-world road corner case dataset for object detection in autonomous driving. arXiv preprint arXiv:2203.07724  (2022)

\bibitem{li2023trackdiffusion}
Li, P., Liu, Z., Chen, K., Hong, L., Zhuge, Y., Yeung, D.Y., Lu, H., Jia, X.: Trackdiffusion: Multi-object tracking data generation via diffusion models. arXiv preprint arXiv:2312.00651  (2023)

\bibitem{li2024automated}
Li, Y., Zhang, W., Chen, K., Liu, Y., Li, P., Gao, R., Hong, L., Tian, M., Zhao, X., Li, Z., et~al.: Automated evaluation of large vision-language models on self-driving corner cases. arXiv preprint arXiv:2404.10595  (2024)

\bibitem{liu2024improved}
Liu, H., Li, C., Li, Y., Lee, Y.J.: Improved baselines with visual instruction tuning. In: CVPR (2024)

\bibitem{liu2023geomerasing}
Liu, Z., Chen, K., Zhang, Y., Han, J., Hong, L., Xu, H., Li, Z., Yeung, D.Y., Kwok, J.: Geom-erasing: Geometry-driven removal of implicit concept in diffusion models. arXiv preprint arXiv:2310.05873  (2023)

\bibitem{liu2024mixture}
Liu, Z., Gou, Y., Chen, K., Hong, L., Gao, J., Mi, F., Zhang, Y., Li, Z., Jiang, X., Liu, Q., et~al.: Mixture of insightful experts (mote): The synergy of thought chains and expert mixtures in self-alignment. arXiv preprint arXiv:2405.00557  (2024)

\bibitem{liu2022task}
Liu, Z., Han, J., Chen, K., Hong, L., Xu, H., Xu, C., Li, Z.: Task-customized self-supervised pre-training with scalable dynamic routing. In: AAAI (2022)

\bibitem{gpt4v_2}
OpenAI: Gpt-4v(ision) technical work and authors. \url{https://openai.com/contributions/gpt-4v} (2023)

\bibitem{rombach2022high}
Rombach, R., Blattmann, A., Lorenz, D., Esser, P., Ommer, B.: High-resolution image synthesis with latent diffusion models. In: CVPR (2022)

\bibitem{wang2024detdiffusion}
Wang, Y., Gao, R., Chen, K., Zhou, K., Cai, Y., Hong, L., Li, Z., Jiang, L., Yeung, D.Y., Xu, Q., Zhang, K.: Detdiffusion: Synergizing generative and perceptive models for enhanced data generation and perception. arXiv preprint arXiv:2403.13304  (2024)

\bibitem{wen2023road}
Wen, L., Yang, X., Fu, D., Wang, X., Cai, P., Li, X., Ma, T., Li, Y., Xu, L., Shang, D., et~al.: On the road with gpt-4v (ision): Early explorations of visual-language model on autonomous driving. arXiv preprint arXiv:2311.05332  (2023)

\bibitem{wu2024unified}
Wu, J., Chung, T.T., Chen, K., Yeung, D.Y.: Unified triplet-level hallucination evaluation for large vision-language models. arXiv preprint arXiv:2410.23114  (2024)

\bibitem{zhili2023task}
Zhili, L., Chen, K., Han, J., Lanqing, H., Xu, H., Li, Z., Kwok, J.: Task-customized masked autoencoder via mixture of cluster-conditional experts. In: ICLR (2023)

\end{thebibliography}
\end{document}